\pdfoutput=1

\documentclass[11pt]{article}

\usepackage{acl}

\usepackage{times}
\usepackage{latexsym}

\usepackage[T1]{fontenc}

\usepackage[utf8]{inputenc}

\usepackage{microtype}
\usepackage{graphicx}

\title{Are ChatGPT and GPT-4 Good Poker Players? - \\ A Pre-Flop Analysis}

\author{Akshat Gupta \\
  UC Berkeley\\
  \texttt{akshat.gupta@berkeley.edu}}

\begin{document}
\maketitle
\begin{abstract}
Since the introduction of ChatGPT and GPT-4, these models have been tested across a large number of tasks. Their adeptness across domains is evident, but their aptitude in playing games, and specifically their aptitude in the realm of poker has remained unexplored. Poker is a game that requires decision making under uncertainty and incomplete information. In this paper, we put ChatGPT and GPT-4 through the poker test and evaluate their poker skills. Our findings reveal that while both models display an advanced understanding of poker, encompassing concepts like the valuation of starting hands, playing positions and other intricacies of game theory optimal (GTO) poker, both ChatGPT and GPT-4 are NOT game theory optimal poker players.

Profitable strategies in poker are evaluated in expectations over large samples. Through a series of experiments, we first discover the characteristics of optimal prompts and model parameters for playing poker with these models. Our observations then unveil the distinct playing personas of the two models. We first conclude that GPT-4 is a more advanced poker player than ChatGPT. This exploration then sheds light on the divergent poker tactics of the two models: ChatGPT's conservativeness juxtaposed against GPT-4's aggression. In poker vernacular, when tasked to play GTO poker, ChatGPT plays like a \textit{nit}, which means that it has a propensity to only engage with premium hands and folds a majority of hands. When subjected to the same directive, GPT-4 plays like a \textit{maniac}, showcasing a loose and aggressive style of play. Both strategies, although relatively advanced, are not game theory optimal. 
\end{abstract}

\section{Introduction}

\begin{figure}
  \centering
  \includegraphics[width=0.40\textwidth]{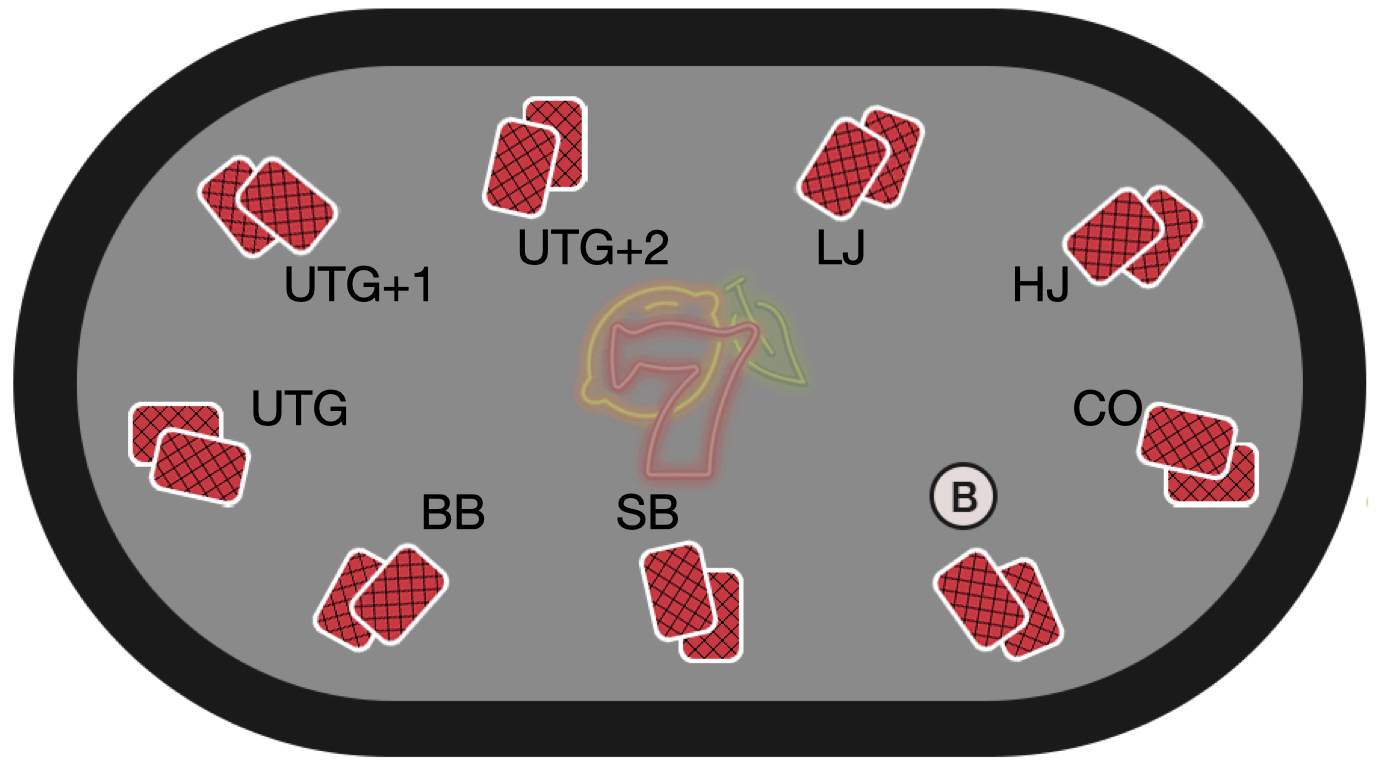}
  \caption{A 9-player poker table with the positions named. The order of action in the pre-flop betting round is from UTG to BB. }
  \label{fig:positions}
\end{figure}
ChatGPT \citep{gpt3.5, chatgpt} and GPT-4 \cite{gpt4} are immensely powerful language models, capable of previously unimaginable tasks. These LLMs go beyond language understanding tasks and are able to do mathematics \cite{frieder2023mathematical} and reasoning \cite{liu2023evaluating}, competitively passing bar exams for becoming a board certified lawyer \citep{bommarito2022gpt, katz2023gpt}, and are able to understand human emotions \cite{elyoseph2023chatgpt} among other things. Playing games require a combination of many such skills, which makes it an interesting setting to test the capabilities of these models. While much attention has been given to analysing the capabilities of large language models (LLMs) on different language understanding and reasoning tasks, evaluating their abilities in game play is currently understudied. Poker is one such complex game that requires a combination of skills including mathematical analysis, reasoning, strategic decision making and understanding human behavior and human psychology. Game theory and exploitative decision making based on opponent behavior are at the heart of the game. In this paper, we put ChatGPT and GPT-4 to the test and evaluate their ability to play the game of poker. 

\begin{figure*}
  \centering
  \includegraphics[width=0.80\textwidth]{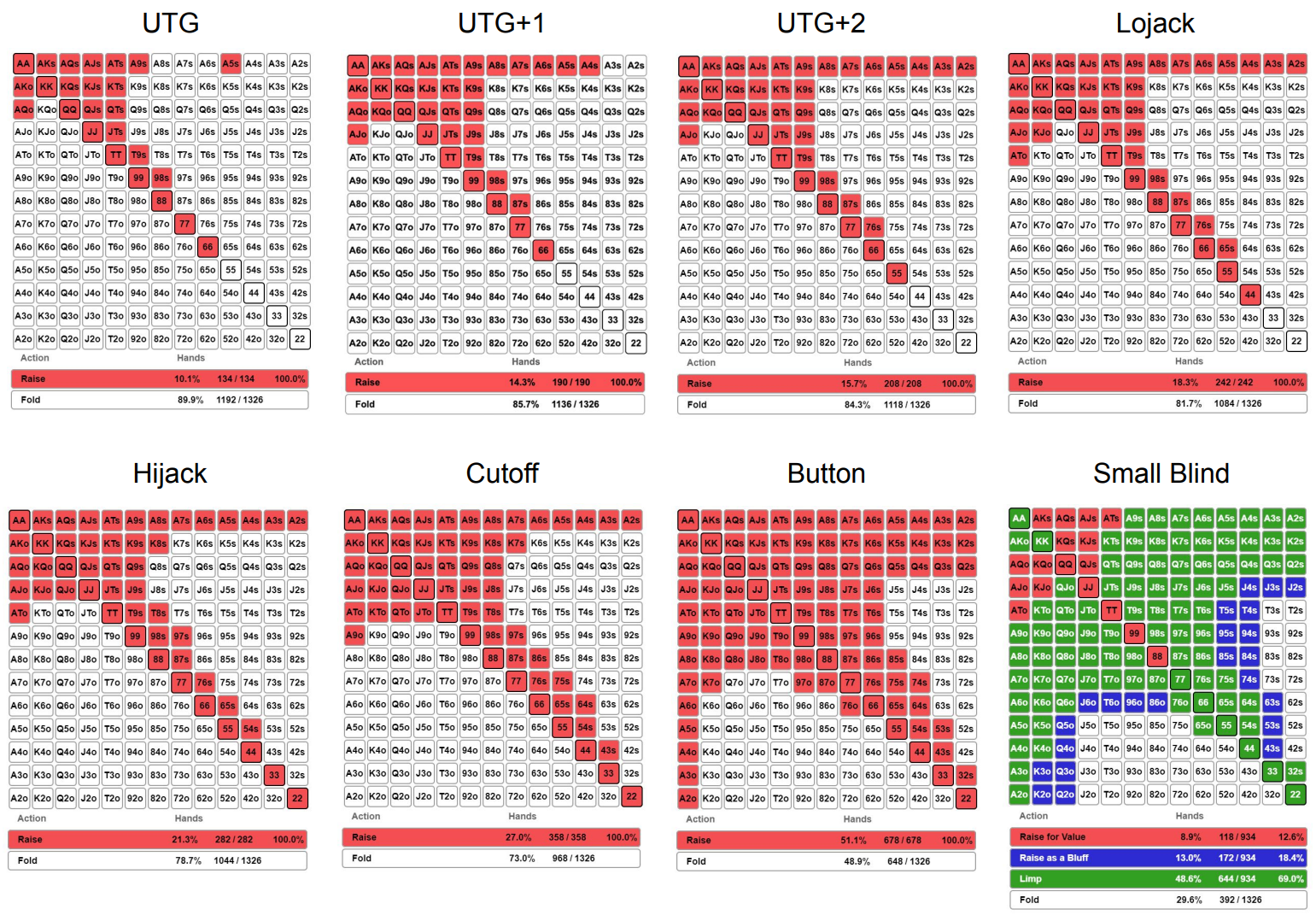}
  \caption{GTO pre-flop strategy in Raise First In spots \cite{rficharts}. The red color shows raised hand, green show limped hands, and white shows folded hands. }
  \label{fig:rfi}
\end{figure*}

Poker is a popular card game that is played in various forms throughout the world. The game consists of multiple rounds, where players receive private information in the form of their own cards, and public information is gradually revealed in the form of shared cards, known as the \textit{community cards}. Players make decisions based on their current hand strength, possible future outcomes, and inferred information about opponents' hands based on their actions and style of play. This makes poker a fascinating instance of \textit{incomplete information game} \cite{harsanyi1995games}. While some efforts towards evaluating LLMs in playing games have recently begun \cite{tsai2023can}\cite{akata2023playing}, it still remains a rather unexplored dimension. To the best of our knowledge, this paper is the first instance of evaluating LLMs at playing incomplete information games. Poker is an extremely technical game using game theory solvers to dictate decision making \citep{gtopoker1,gtopoker2}. 

The most popular variant of poker is called \textit{Texas No-Limit Hold'em} (NLH) that epitomizes the challenge of decision-making under uncertainty and incomplete information. There are four different rounds of betting that happen in Texas NLH after the cards are dealt (appendix \ref{sec:pokerprimer}). In this paper, we study the first round of betting, known as the \textit{pre-flop} round. Decisions in poker are deemed profitable in expectation over a large sample size and cannot be evaluated in isolation. Professional poker players use game theory solvers to come up with profitable strategies in poker, called \textit{game theory optimal} (GTO) strategies. Thus, evaluating poker playing capabilities of LLMs requires an understanding of both a technical understanding of GTO poker and LLMs.

In this paper, we study the pre-flop decisions made by ChatGPT and GPT-4 in a 9-player Texas No-limit Hold'em poker game, when the model is the first to act in the game. The three main factors determining the betting in this round are the private cards held by the player, the position at which the player is playing at, and previous pre-flop bets if any by other players. \textit{Position} in poker refers to the order in which a player acts pre-flop. We go into more detail about positions in poker in a later sections. While LLMs like ChatGPT and GPT-4 were trained on a large amount of internet data, they were not trained to specifically play poker. Thus the ability to play poker, especially GTO poker, cannot be taken for granted. Our paper, to the best of our knowledge, is the first work studying the use of LLMs for playing poker. 

Our findings reveal that both ChatGPT and GPT-4 have an advanced understanding of the game of poker, where they understands concepts like position and starting hand ranges (appendix  \ref{sec:pokerprimer}). \textbf{Yet, both ChatGPT and GPT-4 are not game theory optimal}. ChatGPT is less advanced at playing poker when compared to GPT-4 and tends to play conservatively - folding most hands and only playing with premium hands. GPT-4 on the other hand is an overaggressive player and raises a larger than optimal number of hands.




\section{A Crash Course in No-Limit Hold'em Poker}
In this paper, we ask ChatGPT and GPT-4 to make pre-flop decisions in a 9-player NLH poker game. Figure \ref{fig:positions} shows the pre-flop setting in a 9-player Texas NLH game. The pre-flop round is the first betting round in a NLH game which happens right after the private cards are dealt to the players. In a 9 player scenario, the first two positions to act - the small blind (SB) and big blind (BB) have to put in the chips without seeing their cards. The first person to act voluntarily in the pre-flop scenario is the player after the big blind, called the \textit{under-the-gun} (UTG) player. The players next to act after the UTG player are called UTG+1 (pronounced as under-the-gun-plus-one) and UTG+2. The next positions are usually middle positions. The position to act after UTG+2 is called the \textit{Lojack} (LJ), followed by Hijack (HJ), Cutoff (CO) and the Button (B). Position is a very important factor while making any decision in poker (appendix \ref{sec:pokerprimer}).

In this paper, we study the first step in the pre-flop betting scenario called the \textbf{raise-first-in} (RFI). In this scenario, the player is the first to put chips in the pot. This can happen either if a player is first to act (UTG) or if all players before the current player have decided not to play (have folded). Thus, the players usually choose from one of the following basic actions in the RFI scenario - \textbf{call} or to equal the previous bet (also known as \textit{limp} in pre-flop round), \textbf{raise} or to bet more chips than the bet made by a previous player, or \textbf{fold} or to give up a claim to the and discontinuing in the hand. We refer the reader to appendix \ref{sec:pokerprimer} for a more detailed introduction to the game of poker.

\subsection{Poker Charts}\label{sec:pokercharts}
Strategies in poker are represented using poker charts. Poker charts, also known as starting hand charts, are tools designed to guide players in their decision-making process, especially during the pre-flop stage of a NLH poker game. These charts provide a visual representation of the potential strength of each two-card starting hand, and often suggest an optimal course of action (such as fold, limp, or raise) depending on a player's position at the table. The game-theory-optimal RFI pre-flop charts \cite{rficharts} for different positions are shown in Figure \ref{fig:rfi}. Poker chart to poker players are what periodic tables are to chemists. All poker players remember many such poker charts by heart to avoid needing computational solvers (which are never allowed in live games) and yet be able to make game-theory optimal decisions.

These poker charts are arranged in a 13 by 13 matrix. This is a compressed representation of 1326 possible starting hands that a player can have (poker is played with a deck of 52 cards, with the cards divided into 4 suits of 13 cards each). The suited cards form the upper diagonal matrix of the poker charts, depicted by 's', and the unsuited starting cards are denoted by 'o' in the lower diagonal matrix of the poker charts (`o' stands for off-suite). The diagonal elements contain two starting cards with the same number, called \textit{pocket pairs}. For a more comprehensive explanation of poker charts, we refer the reader to appendix \ref{sec:pokercharts}.

\subsection{GTO Pre-flop Strategy}\label{sec:gtopoker}
Poker charts shown in figure \ref{fig:rfi} contains game theory optimal RFI pre-flop decisions for different positions. The RFI condition assumes that all players before the current player have folded or the current player is the first to act. Some very easy to observe patterns for GTO pre-flop play is that UTG player folds most hands. As can be seen in figure \ref{fig:rfi}, UTG player folds approximately 90\% of their hands, whereas a player on the button position only folds 50\% of their hands. This shows that one should only play a very restricted set of hands from early positions and can increase their range of cards from later positions. The second immediate observation is that GTO poker does not recommend that you limp (except at small blind position). That means that in general, the GTO strategy pre-flop is to either raise, or fold. 

The minimum bet in the game is a pre-decided quantity that is usually fixed for the duration of the entire game (considering standard cash games). Coincidentally, the amount of the minimum bet is also called \textbf{big blind} (BB). For example, if the minimum amount you can bet at a table is 3\$, then 1 BB = 3\$.The GTO raise amount is usually considered to be 3 BB, which in this example would be equal to 9\$. To clarify for the readers, big blind is a term used both for a position in poker and the minimum amount that can be bet in a game, and is disambiguated by the context. 

\begin{figure*}[h]
  \centering
  \includegraphics[width=0.88\textwidth]{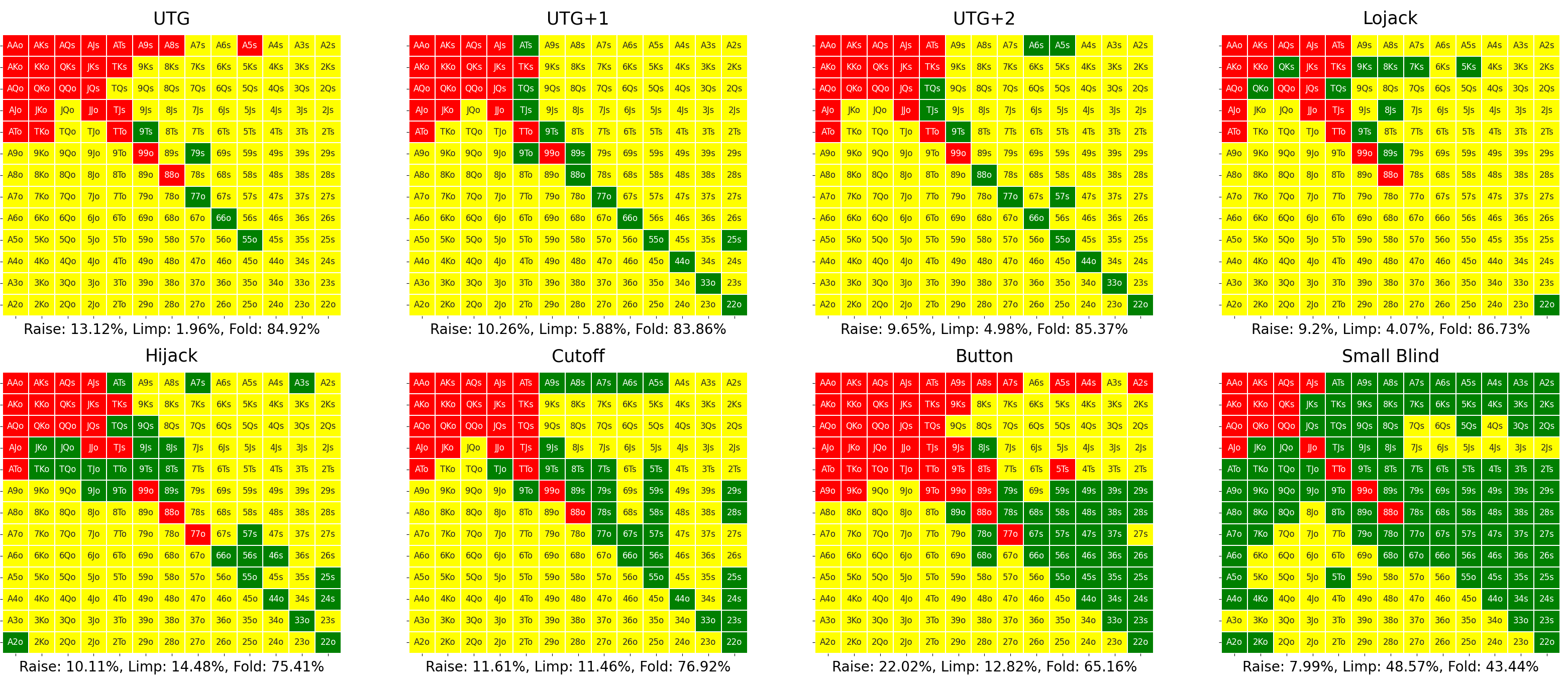}
  \caption{ChatGPT's pre-flop strategy in Raise First in spots. The red color hands show raised hands, green color shows limped hands and yellow shows folded hands. This decision matrix is for \textbf{short}-type and \textbf{ranked} user prompt.}
  \label{fig:rfi_basic_prompt_short}
\end{figure*}

\begin{figure*}
  \centering
  \includegraphics[width=0.88\textwidth]{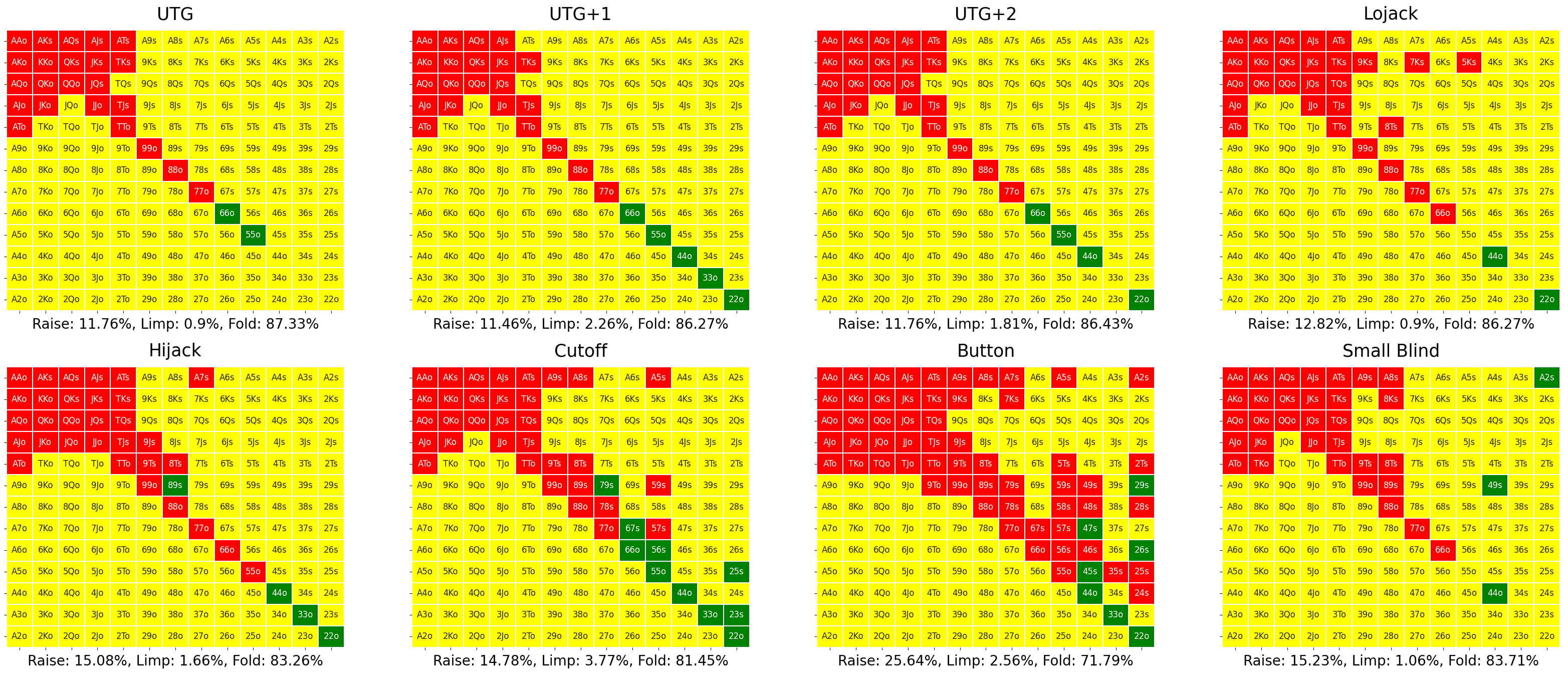}
  \caption{ChatGPT's pre-flop strategy in Raise-First-in using the \textbf{short} user prompt and \textbf{ranked} order of card presentation. Here, ChatGPT is specifically asked to play \textbf{GTO poker}.}
  \label{fig:rfi_chatgpt_gto_ranked_short}
\end{figure*}

\section{ChatGPT Playing Poker}
We now move on to first have ChatGPT (gpt-3.5-turbo) play poker. We specifically analyse ChatGPT's decision making in the RFI pre-flop step. The first step in this process is to choose the right prompts to get ChatGPT to play poker. We experiment with different types of prompts. In the sections that follow, we go through the different prompts tried, our rationale for them and what we learnt about how to play poker with ChatGPT.  

\begin{figure*}
  \centering
  \includegraphics[width=0.85\textwidth]{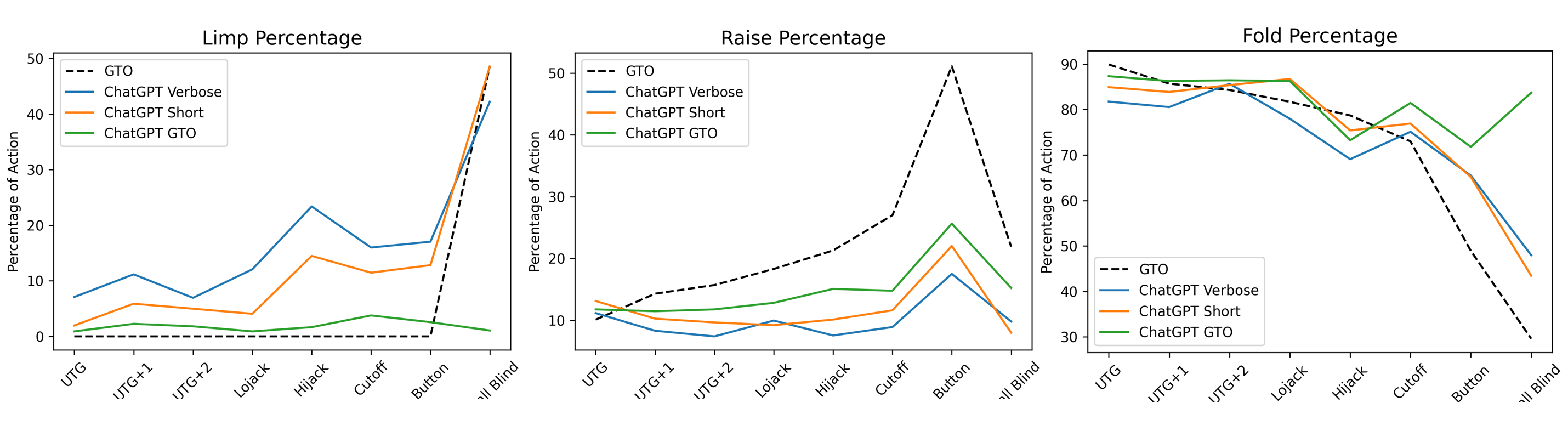}
  \caption{ChatGPT action percentage as a function of position. Action options that ChatGPT has are raise, fold and limp. The positions on the x-axis are ordered in order of action on the table.}
  \label{fig:chatgpt_action_percentage}
\end{figure*}

\begin{figure*}
  \centering
  \includegraphics[width=0.85\textwidth]{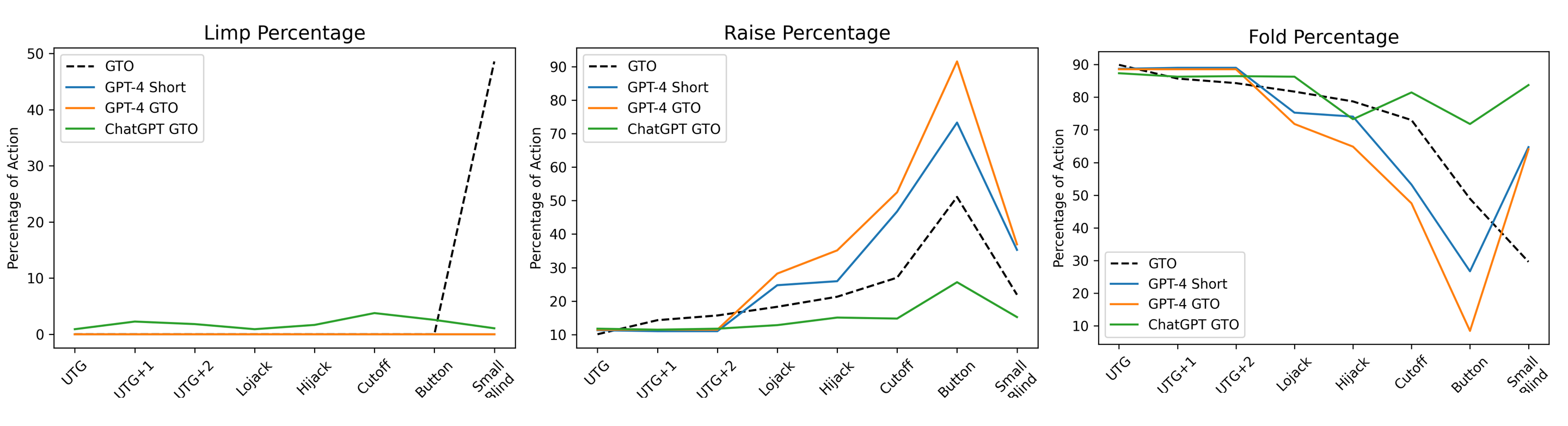}
  \caption{GPT-4 action percentage as a function of position. `GPT-4 short' is the standard GPT-4 model with \textit{short}-type and ranked format prompting. }
  \label{fig:gpt4_action_percentage}
\end{figure*}

\subsection{Basic System Prompt}
After a lot of experimentation based on analysing model responses, we use the following system prompt to describe the RFI pre-flop setting to ChatGPT:

{\small
\begin{quote}
\textit{You are playing a 9 player Texas No-limit Holdem poker game. You will be provided with your position at the table and the hand you're holding. Please provide your pre-flop decision.\\\\ Assume you are the first to act and everyone before you has folded, thus your decisions can be one of fold, raise or limp. If you are placing a bet, please specify your best size in terms of big blinds.\\\\Provide your decision without any explanation in the following format: DECISION(Raise, Fold, Limp), N BB (if placing a bet, replace N by bet amount)}
\end{quote}
}

The above system prompt was carefully selected after trying many different prompts. The criteria for selecting the prompt was firstly, being able to appropriately describe the scenario of the game, and secondly, having ChatGPT generate output in the desired format. The first line in the prompt describes the game ChatGPT is playing, which is a 9-player Texas no-limit Holdem poker. ChatGPT is then provided with information about the RFI scenario, where it is asked to assume that everyone before it has folded and it is the first to act in the hand. To further aid ChatGPT in making decisions, we specify the options it has in such a scenario, which is either to fold, raise or limp. Finally, ChatGPT is asked to provide its decision in a specific format as described in the prompt.

\subsubsection{User Prompts}

The information about the cards and positions is provided as a user prompt. We experimented with two ways of providing the user prompt. In the first settings (called `verbose'), we expand the name of cards and the \textit{suit} information. In the second type of user prompt (called `short'), we shorten the card names and the suit information to one letter. An example user prompt for both settings is shown below:

\begin{quote}
\begin{itemize}
    \item Verbose : \textit{UTG A,K suited}
    \item Short : \textit{UTG AKs}
\end{itemize}
    
\end{quote}

Our rationale for using the two types of user prompts was to describe the private cards as clearly as possible. The \textit{verbose}-type user prompt describes the current hands more elaborately, and we expect the model to perform better with this user prompt. The \textit{short}-type prompt on the other hand is a more concise representation of the private card information, although it is the standard way of talking about hands in the poker community. All poker charts available online, including the one shown in figure \ref{fig:rfi}, use this notation. The `s' stands for \textit{suited} hands whereas the `o' stands for \textit{offsuit} hands. 

Another variable in choosing the correct prompt was the order in which the cards were presented to ChatGPT. One possible way to present starting hands was to always present the highest ranked of the two cards (Ace being the highest) first and the smaller ranked card second. Thus, if the private cards for a player are Ace and King, it will be written as \textit{AK} in the \textit{ranked} prompts and as \textit{KA} in the \textit{unranked} prompts. We will discuss the effects of these different kinds of prompting in a later section.

\subsection{Analysing ChatGPT's Decision Matrix}
Based on the above system prompt and two types of user prompts, we try to recreate the RFI pre-flop decision charts as shown in figure \ref{fig:rfi} for ChatGPT. The aim is to understand ChatGPT's decision making in this situation and consequently its adeptness in playing poker. With the given system and user prompts, we ask ChatGPT to make a decision for each hand combination on the pre-flop charts at every position. Each positions has 169 possible hands (13 times 13), and there are 8 positions in total, leading to 1352 unique queries. We perform these experiments for three values of temperature (0.2, 0.7, 1.0) and two values of top-p (0.95, 1), thus leading to 6 experiments for each of the ways of prompting. We prompt the model 10 times for each hand, and choose the most common pre-flop decision made by the model to counteract errors due to sampling based generation. We find that temperature = $0.2$ and top-p = $0.95$ produces the most robust results which are closest to GTO results. In this section, we only show results for temperature = $0.2$ and top-p = $0.95$. The chosen values of temperature lead to generation of most probable answer \citep{curios, gpt2, gpt3}, \textbf{thus showing that most probable answer is more GTO than sampling based generation for poker}. 

ChatGPT's pre-flop decision matrices for the \textit{short} and \textit{ranked} prompt are shown in figure \ref{fig:rfi_basic_prompt_short}. Since this is our first look at ChatGPT playing poker, it is useful to highlight some non-trivial and yet fundamental observations about ChatGPT playing poker. We want to remind the readers that ChatGPT is not trained to play poker, but just to predict the next word from a huge corpus of internet text \citep{gpt1, gpt2, gpt3}. The internet is full of information about poker, including multiple lessons and poker charts describing game theory optimal (GTO) poker, and a lot of discussions about hands and how to play them on online forums. Any knowledge that these models have about the game of poker would be implicitly learnt through these sources. With the vast amount of knowledge that these models have, its expected for these models to know about the game of poker; but being able to play the game optimally cannot be assumed.

\textbf{Observation 1 - ChatGPT Understands Poker at and advanced level.} While ChatGPT's proficiency at poker is debatable, there is no doubt that ChatGPT understands poker at a fundamental level. An easy way to see this is to look at the poker charts produced by ChatGPT (figure \ref{fig:rfi_basic_prompt_short}). At every position, ChatGPT always raises with pocket Aces (AA), which is the best starting hand in poker, and always folds 27 offsuit (27o), which is the worst starting hand in poker. It follows similar patterns of raising and folding with the few other top and worst starting hands in poker. While understanding of rules and relative winning potential of starting hands comes at early stages of playing, understanding the importance of position is a more advanced concept. As discussed earlier, the GTO way to play pre-flop is to play fewer hands from earlier positions, and to play a larger percentage of hands from later positions. We can clearly see this pattern being followed in the ChatGPT pre-flop charts, as show in figures \ref{fig:rfi_basic_prompt_short}. We can also see this in figure \ref{fig:chatgpt_action_percentage}, where the fold percentage decreases as ChatGPT plays from later positions.

\textbf{Observation 2 - There is a \textit{Correct} way to ask ChatGPT to play Poker:} There is a right way and a wrong way to ask ChatGPT to make pre-flop decisions. The pre-flop cards should be provided in a \textit{short}-type writing, like AKo or AKs rather than writing their more verbose forms. Similarly, writing the higher ranked card first is even more crucial to get more GTO-like decisions. For example, asking ChatGPT to predict A4s vs 4As will lead to completely different decision matrices. This can be seen in figure \ref{fig:rfi_basic_prompt_verbose} and figure  \ref{fig:rfi_basic_prompt_short_unranked}. While we see a minor drop in performance with \textit{verbose} prompts (figure \ref{fig:rfi_basic_prompt_verbose}), \textit{unranked} prompts lead to extremeley non-GTO and erratic style of playing (figure \ref{fig:rfi_basic_prompt_short_unranked}), where the model folds premium hands like AQs. 

\begin{figure*}
  \centering
  \includegraphics[width=0.85\textwidth]{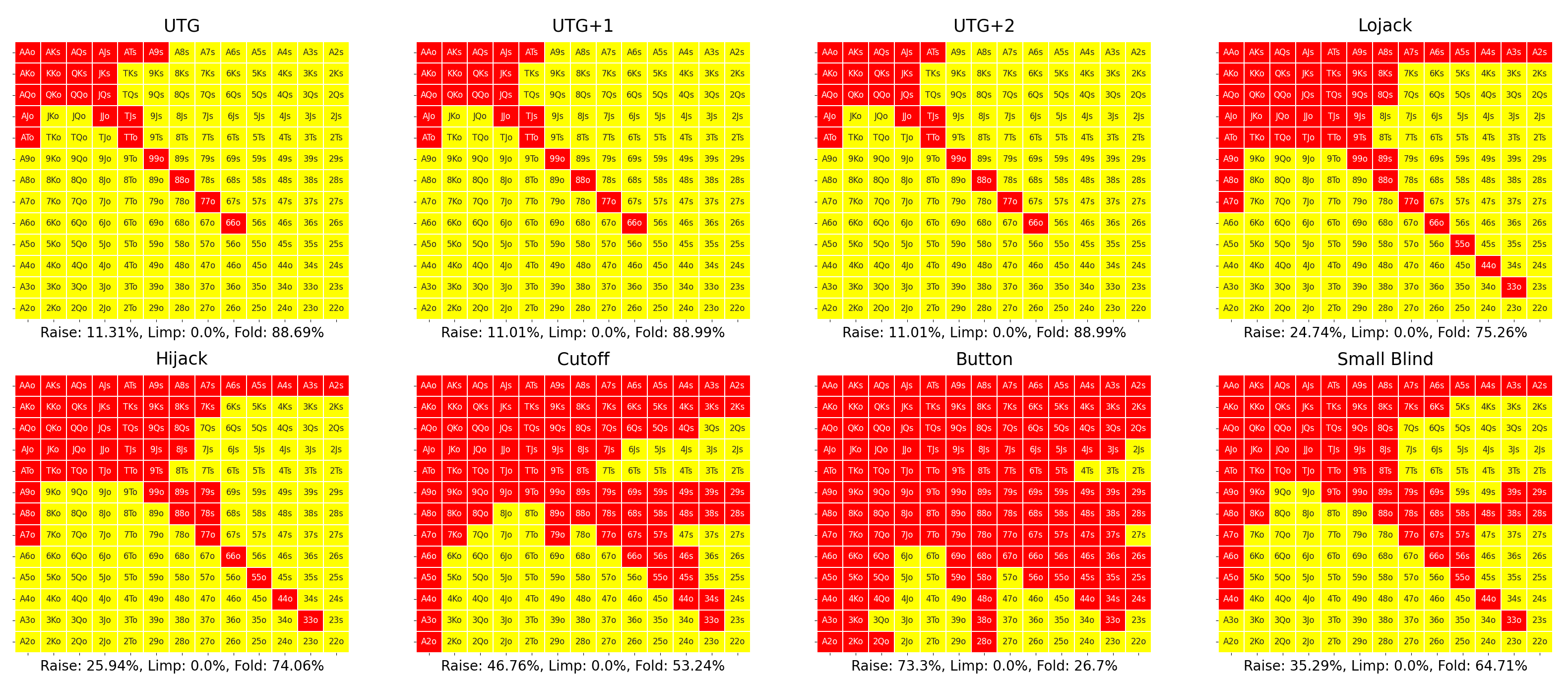}
  \caption{GPT-4's pre-flop strategy in Raise First in spots. This decision matrix is for \textbf{short} user prompt, with the \textbf{unranked} order of card presentation. }
  \label{fig:rfi_gpt4_basic_prompt}
\end{figure*}

\subsection{GTO System Prompt}
The above analysis shows ChatGPT has a deeper understanding of poker than a beginner. If we look at the decision matrices of ChatGPT, we do find hints of a game theory optimal decision making. Since we're comparing ChatGPT's game with a GTO game, why not ask ChatGPT to play GTO poker? By doing this, we will also be able to find if ChatGPT knows how to play GTO poker and change its strategy when specifically asked to play GTO poker. To do this, we change the previous system prompt to explicitly ask ChatGPT to make GTO decisions. To do this, we modify the system prompt as follows:

{\small
\begin{quote}
\textit{You are playing a 9 player Texas No-limit Holdem poker game. You will be provided with your position at the table and the hand you're holding. Please provide your pre-flop decision \textcolor{red}{based on game theory optimal (GTO) poker}.\\\\ Assume you are the first to act and everyone before you has folded, thus your decisions can be one of fold, raise or limp. If you are placing a bet, please specify your best size in terms of big blinds.\\\\Provide your decision without any explanation in the following format: DECISION(Raise, Fold, Limp), N BB (if placing a bet, replace N by bet amount)}
\end{quote}
}

The above prompt follows the same template as before with an additional line asking ChatGPT to make GTO decisions. We continue to use temperature = $0.2$ and top-p = $0.95$. We also use our learnings from the previous analysis, and use the `correct' way to ask ChatGPT to play poker. That is, we provide hand information in a \textit{short}-type user prompt where the hands presented are ranked.

\subsubsection{ChatGPT GTO Decision Matrix Analysis}
ChatGPT decision matrix when asked to make GTO pre-flop RFI decisions is show in Figure \ref{fig:rfi_chatgpt_gto_ranked_short}. One immediate thing to observe is that the number of limps goes down by a significant amount when ChatGPT is asked to make GTO decisions. This can also be seen in Figure \ref{fig:chatgpt_action_percentage}, where the limp percentage of ChatGPT's GTO decisions is very close to zero. The number of hands ChatGPT raises also increases, which is also seen in Figure \ref{fig:chatgpt_action_percentage}.

\textbf{Observation 3 - ChatGPT Understands GTO Poker, although its not GTO.} As we ask ChatGPT to make game theory optimal decisions, the kinds of decisions made by ChatGPT start to resemble GTO poker decisions a lot more. ChatGPT almost halves its limping range and raises a lot more hands, thus becoming more aggressive player than when not prompted to make GTO decisions. Being aggressive is always more profitable in poker than limping passively, although ChatGPT still seems to have a few limps in its range. Additionally, ChatGPT raises with some weird hands, like K7s from the Lojack positions and plays and raises way fewer hands, especially from the button, thus deviating from GTO poker. Ideally, we want all three playing curves of ChatGPT in figure \ref{fig:chatgpt_action_percentage} to be as close to GTO as possible (black dotted line). While it gets closer to GTO for limping by folding junk hands, it is not true for the action of raising. This style of play indicates that \textbf{ChatGPT might fall into a category of a player called \textit{Nit}}. \textit{Nit} is a term used in poker terminology to describe players who are very tight and conservative. Being a \textit{Nit} is not a profitable style of playing poker since this playing style can lead to missed opportunities, predictability, and vulnerability to aggressive opponents. 

\begin{figure*}
  \centering
  \includegraphics[width=0.85\textwidth]{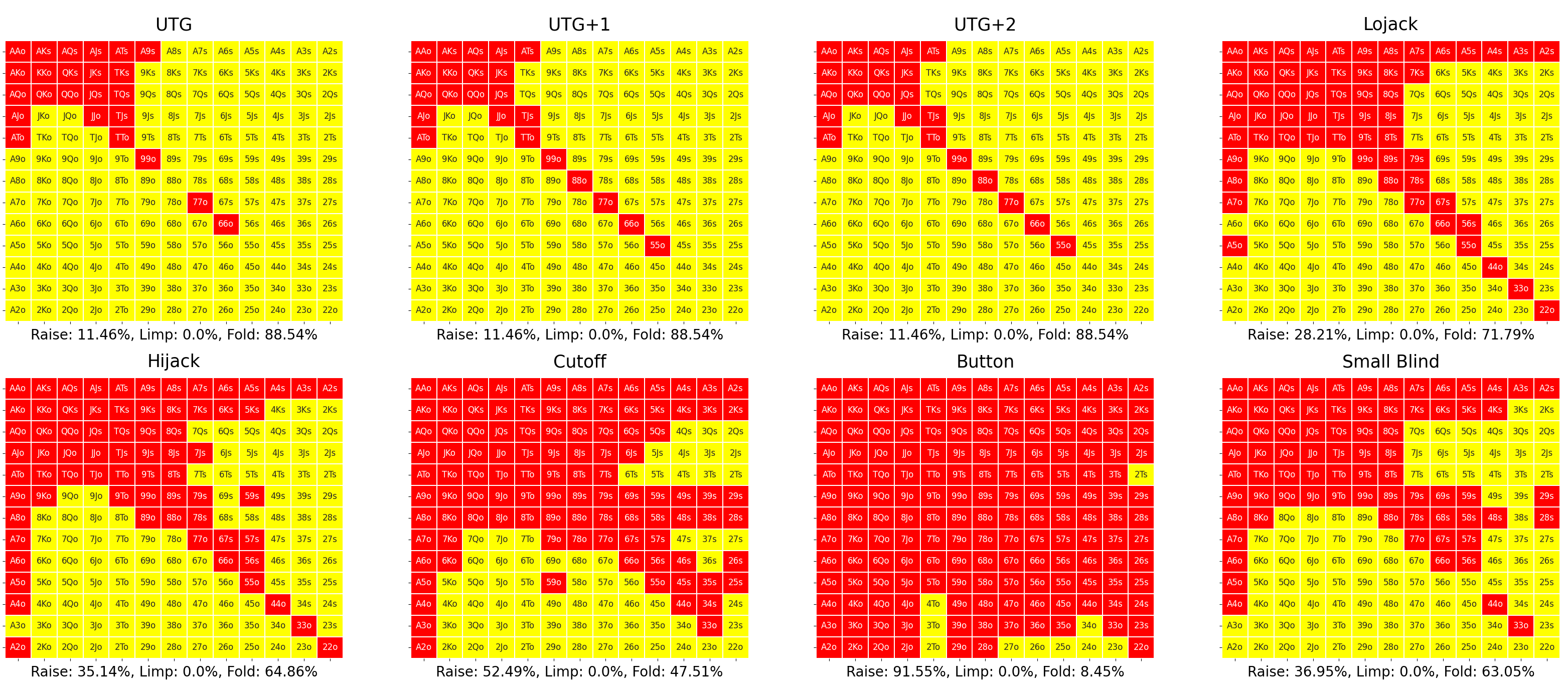}
  \caption{GPT-4's pre-flop strategy in Raise First in spots for \textbf{short} user prompt, with the \textbf{unranked} order of card presentation. Here, GPT-4 is specifically asked to play \textbf{GTO poker}}
  \label{fig:rfi_gpt4_gto}
\end{figure*}

\section{GPT-4 Playing Poker}
Next we have GPT-4 play poker and analyse it's style of play. We use the two system prompts described in the above sections - basic prompt and GTO prompt. We also use the \textit{short}-type user prompt to provide hand information in a ranked manner where the higher ranked card is provided first. We also use a temperature = 0.2 and top-p = 0.95. Because of the higher querying cost, we query GPT-4 five times for each hand and present the majority decision on the decision matrices. 

The RFI pre-flop decision matrices of GPT-4 are shown in figures \ref{fig:rfi_gpt4_basic_prompt} and \ref{fig:rfi_gpt4_gto}. We see that GPT-4 too has a deep understanding of the game of poker and more advanced poker concepts like position. One striking observation in both basic and GTO prompts is that there are absolutely no limps in the pre-flop range of GPT-4. We also see that the decision charts of basic and GTO prompt are very similar except at later positions like the Button or Small Blind. This indicates that GPT-4 is naturally a more GTO player than ChatGPT, as it limps less, raises more and plays a larger number of hands compared to GTO ChatGPT without even being prompted to be game theory optimal.

\textbf{Observation 4 - GPT-4 is more aggressive than ChatGPT.} The first reason for this observation is the fact that GPT-4 never limps. This itself makes it a more aggressive player as compared to ChatGPT. Additionally, as seen in Figure \ref{fig:gpt4_action_percentage}, GPT-4 even with the basic prompt raises way more hands than GTO ChatGPT. GPT-4 also plays a larger number of hands compared to GTO ChatGPT, and gets even more aggressive when asked to be GTO, as seen in Figure \ref{fig:gpt4_action_percentage}. Not limping, raising when being involved in the pot and playing a larger number of hands are signs of a more aggressive player.

\textbf{Observation 5 - GPT-4 still not GTO.} GPT-4 starts of playing close to GTO from early positions, although from later positions, it becomes really loose and aggressive. Starting from the Lojack position, GPT-4 raises more hands than game theory optimal, ending up raising 90\% of the hands from the Button when asked to be GTO. \textbf{GPT-4's aggression is almost similar to a player type called \textit{maniac} in poker}. A \textit{maniac} in poker refers to a highly aggressive player who raises with a wide range of hands pre-flop, avoids limping, and exhibits unpredictable and unconventional play. While being aggressive in poker is generally profitable, being overly aggressive is usually not a profitable strategy because it can lead to high variance in returns, unpredictable play, and potential losses due to overly aggressive and non-strategic betting. Although GPT-4 might be a more advanced player when compared to ChatGPT, it is still not GTO.

\section{Conclusion}
We studied the ability of ChatGPT and GPT-4 to play the game of poker. We do this using by testing their playing style against game theory optimal way of playing poker. Through our experiments, we find that both ChatGPT and GPT-4 have an advanced understanding of the game of poker. They don't just understand the rules of poker but also the intricacies of better and worse starting hands. Both models also understand the concept of position in poker and tend to play differently based on their position on the poker table. Yet, both those models are not game theory optimal


Both models also seem to have an understanding of what playing game theory optimal poker means, although both models have a different reaction when asked to play GTO. When asked to play GTO, ChatGPT removes a large number of limps and raises a larger portion of hands that it plays, thus becoming more aggressive. Additionally, ChatGPT also starts playing fewer number of hands. Thus, for ChatGPT, playing GTO means to become tighter and more aggressive. GPT-4 is already an aggressive player and in order to become GTO, it neeeds to become tighter. Yet, when asked to play GTO, GPT-4 raises an even larger number of hands, mostly from the later positions. Thus for GPT-4, playing GTO means to be playing more aggressively, which is the opposite of what it needs to do. This almost points to the fact that these models are unaware of their own playing tendencies. 

It is interesting to see that these two models have exactly opposite tendencies which prevent them from being game theory optimal. ChatGPT plays a tight and conservative game, and limps some of its weaker hands. Thus, ChatGPT is not game theory optimal because it is less aggressive. GPT-4 on the other hand does not limp at all and raises all the hands it plays, yet it seems to be raising a lot more hands than necessary. Thus, GPT-4 is not game theory optimal because it is overly aggressive.

\section{Limitations}
As pointed out in different points of the paper, we analyse ChatGPT and GPT-4's decisions when playing poker in the raise-first-in scenario of the pre-flop betting round, which is the first decision making point in poker. Hence, our analysis of the poker playing abilities of these LLMs are just based on the first decision point in poker, which is the easiest to evaluate both computationally and conceptually. Just doing this analysis required as to make hundreds of thousands queries to OpenAI APIs. 

Pre-flop scenarios are also the simplest to get right when playing GTO poker, so there is absolutely no reason to expect ChatGPT or GPT-4 to become more GTO in later decisions points of the game. In fact, current deviation from GTO poker suggests that these models will only get worse, since all future decision making points in poker are dependent on the previous decision and the errors propagate.


\bibliography{anthology,custom}

\newpage
\appendix

\section{Appendix}
\subsection{Poker Primer}\label{sec:pokerprimer}
The most popular variant of poker is called \textit{Texas No-Limit Hold'em} (NLH) that epitomizes the challenge of decision-making under uncertainty and incomplete information. In this variant, each player is dealt two private cards, and five community cards are dealt (not all at once) face-up on the board visible to everyone. There are four different rounds of betting that happen in Texas NLH as explained below:

\begin{itemize}
    \item \textbf{Pre-Flop} : This round of betting happens right after the players see their two private cards. Players have to choose from a few choices in poker including betting, folding, raising etc. The different kinds of decisions made in NLH poker are discussed later in the paper.
    \item \textbf{Post-Flop} : Once the pre-flop betting round finishes, three community cards are dealt face-up at the table visible to everyone. The event of dealing of the first three cards in NLH is called the \textit{flop}. One round of betting happens right after the first three community cards are dealt, called the post-flop betting round. 
    \item \textbf{Post-Turn} : After the post-flop betting round, a fourth community card is dealt, again face-up and visible to everyone, called the \textit{turn} card. Another round of betting happens after the dealing of the turn card.
    \item \textbf{Post-River} : After post-turn betting round, the fifth and final community card is dealt, also face-up and visible to everyone. The final card is called the \textbf{river} card which is followed by the final round of betting. 
\end{itemize}

\subsubsection{The Pre-Flop Setting in No-Limit Hold'em Poker}
In this paper, we ask ChatGPT and GPT-4 to make pre-flop decisions in a 9-player NLH poker game. Figure \ref{fig:positions} shows the pre-flop setting in a 9-player Texas NLH game. The pre-flop round is the first betting round in the game which happens right after the private cards are dealt to the players. As part of standard poker rules, two players have to put a specific amount of money on the table without seeing their private cards. These players are called the \textit{blinds}. The player called the \textit{small blind} puts in half of the minimum bet that can be made in the game. The player called the \textit{big blind} has to put an amount equal to the minimum bet that can be made in the game. The minimum bet in the game is a pre-decided quantity that is usually fixed for the duration of the entire game (considering standard cash games). Coincidentally, the amount of the minimum bet is also called \textbf{big blind} (BB). For example, if the minimum amount you can bet at a table is 3\$, then 1 BB = 3\$. A common starting bet in poker is 3 BB, which in this example would be equal to 9\$. To clarify for the readers, big blind is a term used both for a position in poker and the minimum amount that can be bet in a game, and is disambiguated by the context. We will be using the abbreviation \textit{BB} to specifically refer to the amount of minimum bet and will never refer to the position by these abbreviations.

Since the small blind and big blind have to put in the chips without seeing their cards, the first person to act in the pre-flop scenario is the player after the big blind called the \textit{under-the-gun} (UTG) player. The players next to act after the UTG player are called UTG+1 (pronounced as under-the-gun-plus-one) and UTG+2. The next positions are usually middle positions. In a 9 player scenario, the position to act after UTG+2 is called the \textit{Lojack} (LJ), followed by Hijack (HJ), Cutoff (CO) and the Button (B). Position is a very important factor while making any decision in poker. These positions can be seen in figure \ref{fig:positions}.

In this paper, we study the first step in the pre-flop betting scenario called the \textbf{raise-first-in} (RFI). In this scenario, the player is the first to put chips in the pot. This can happen either if a player is first to act (UTG) or if all players before the current player have decided not to play (have folded). Thus, the players usually choose from one of the following basic actions in the RFI scenario:

\begin{itemize}
    \item \textbf{Bet}: The act of placing a wager into the pot during a betting round. In poker, the term \textit{bet} is specifically referred to the scenario when no previous player has wagered chips in the ongoing round, then the first player to wager chips is said to have \textit{bet}. UTG player is the first to bet in the pre-flop round since the blinds do not place a voluntary bet. 
    \item \textbf{Call}: The action of matching the bet made by a previous player is called a \textit{call}. In the specific scenario of RFI, if a player matches the 1BB bet made by the big blind, then this calling action is referred to as a \textit{limp}. 
    \item \textbf{Raise}: Betting more chips than the bet made by a previous player called a \textit{raise}. If a player raises, other players must either match the increased bet (call), raise it further, or fold.
    \item \textbf{Fold}: Choosing not to match a bet or a raise and therefore giving up any claim on the pot is called \textit{folding}. A player who folds is out of action for the remainder of the hand.
\end{itemize}

In the RFI pre-flop setting, the only possible actions a player can take are Limp, Raise or Fold. 

\subsubsection{Poker Charts}\label{sec:pokercharts}
Strategies in poker are represented using poker charts. Poker charts, also known as starting hand charts, are tools designed to guide players in their decision-making process, especially during the pre-flop stage of a NLH poker game. These charts provide a visual representation of the potential strength of each two-card starting hand, and often suggest an optimal course of action (such as fold, limp, or raise) depending on a player's position at the table. Position is crucial in poker as it determines the order of play, and having later position often provides a strategic advantage. The game-theory-optimal RFI pre-flop charts for different positions are shown in Figure \ref{fig:rfi}. Note that there is a different decision matrix for every starting position, and in general, the earlier the position, the fewer hands are played. Poker chart to poker players are what periodic tables are to chemists. All poker players remember many such poker charts by heart to avoid needing computational solvers (which are never allowed in live games) and yet make game-theory optimal decisions.

These poker charts are arranged in a square of 13 by 13. This is a compressed representation of 1326 possible starting hands that a player can have (poker is played with a deck of 52 cards, with the cards divided into 4 suits of 13 cards each). Each starting hand can either be \textit{suited}, which means that both cards belong to the same suit. For the purposes of a starting hand, Ace-King of Diamonds is equivalent to Ace-King of Hearts. The only relevant information here is that the cards have the same suit. Similarly, one combination of suits, example Spade-Hearts, is in no way different from another combination of suits like Club-Diamond. Hence, the only relevant information is that both cards have different suits. Therefore, apart from the numbers of the two cards, the only other relevant information that needs to be considered for making pre-flop decisions is whether the pair of cards are suited or unsuited. The suited cards form the upper diagonal matrix of the poker charts, depicted by 's', and the unsuited starting cards are denoted by 'o' in the lower diagonal matrix of the poker charts, with the diagonal elements containing two cards with the same number, called \textit{pocket pairs}. 

\begin{figure*}[h]
  \centering
  \includegraphics[width=1\textwidth]{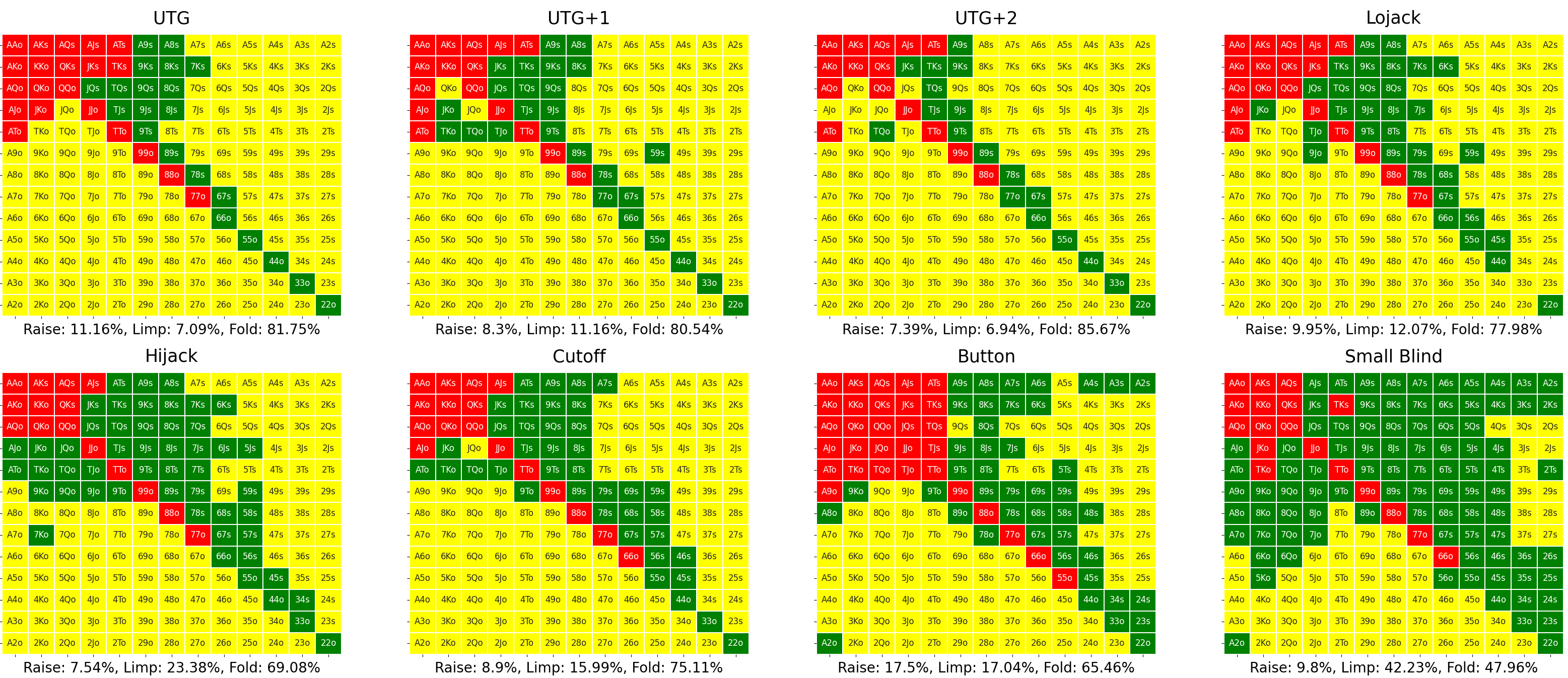}
  \caption{ChatGPT's pre-flop strategy in Raise First in spots. The red color hands show raised hands, green color shows limped hands and yellow shows folded hands. This decision matrix is for \textbf{verbose} user prompt.}
  \label{fig:rfi_basic_prompt_verbose}
\end{figure*}

\begin{figure*}
  \centering
  \includegraphics[width=1\textwidth]{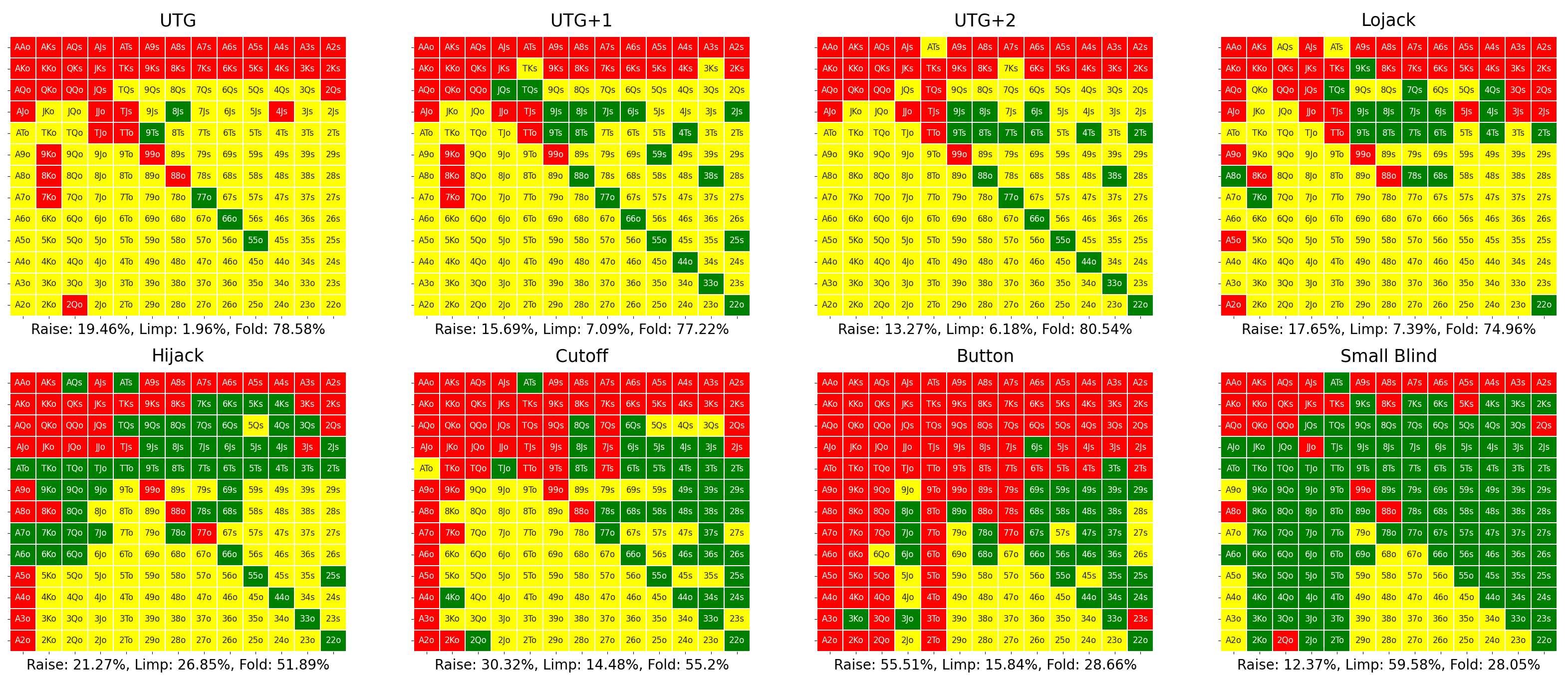}
  \caption{ChatGPT's pre-flop strategy in Raise First in spots. The red color hands show raised hands, green color shows limped hands and yellow shows folded hands. This decision matrix is for \textbf{short} user prompt, with the \textbf{unranked} order of card presentation. }
  \label{fig:rfi_basic_prompt_short_unranked}
\end{figure*}

\end{document}